\documentclass[10pt,twocolumn,letterpaper]{article}

\usepackage{iccv}      

\definecolor{iccvblue}{rgb}{0.21,0.49,0.74}
\usepackage[pagebackref,breaklinks,colorlinks,allcolors=iccvblue]{hyperref}

\def\paperID{12744} 
\def\confName{ICCV}
\def\confYear{2025}
\usepackage{times}  
\usepackage{amsmath}
\usepackage{amssymb}
\usepackage{helvet}  
\usepackage{courier}  
\usepackage[hyphens]{url}  
\usepackage{subcaption}
\usepackage{natbib}  
\usepackage{caption} 
\usepackage{algorithm}
\usepackage{algorithmic}
\usepackage{soul}
 \usepackage{multirow}
\usepackage{xspace}
\usepackage{bbding}
\usepackage{pifont}

\newcommand{\name}{\texttt{BadSFL}\xspace}

\newcommand{\one}{\raisebox{-0.6mm}{\large{\ding{172}}}}
\newcommand{\two}{\raisebox{-0.6mm}{\large{\ding{173}}}}
\newcommand{\three}{\raisebox{-0.6mm}{\large{\ding{174}}}}

\newenvironment{packeditemize}{
\begin{list}{$\bullet$}{
\setlength{\labelwidth}{6pt}
\setlength{\itemsep}{0pt}
\setlength{\leftmargin}{\labelwidth}
\addtolength{\leftmargin}{\labelsep}
\setlength{\parindent}{0pt}
\setlength{\listparindent}{\parindent}
\setlength{\parsep}{0pt}
\setlength{\topsep}{3pt}}}{\end{list}}
\usepackage{balance}

\usepackage{xcolor}

\usepackage{newfloat}
\usepackage{listings}

\title{Mind the Cost of Scaffold! \\Benign Clients May Even Become Accomplices of Backdoor Attack}

\author{%
  Xingshuo Han\textsuperscript{1}, Xuanye Zhang\textsuperscript{1}, Xiang Lan\textsuperscript{2}, Haozhao Wang\textsuperscript{3*}, Shengmin Xu\textsuperscript{2}, Shen Ren\textsuperscript{4},\\
   Jason Zeng\textsuperscript{5}, Ming Wu\textsuperscript{5}, Michael Heinrich\textsuperscript{5}, Tianwei Zhang\textsuperscript{1}\\
  \small\textsuperscript{1}{Nanyang Technological University},\textsuperscript{2}{Fujian Normal University}, \textsuperscript{3}{Huazhong University of Science and Technology}\\
  \small\textsuperscript{4}{Continental Automotive Singapore}, \textsuperscript{5}{Zero Gravity Labs}, \textsuperscript{*}{Corresponding} \\
   \small\textit{\{xingshuo001, C200212, tianwei.zhang\}@ntu.edu.sg},\textit{\{jason, ming, michael\}@0g.ai}\\
   \small\textit{\{lanxiang0113, smxu1989\}@gmail.com}, \textit{hz\_wang@hust.edu.cn, shen@shenren.org} \\
}

\begin{document}

\maketitle

\begin{abstract}
By using a control variate to calibrate the local gradient of each client, Scaffold has been widely known as a powerful solution to mitigate the impact of data heterogeneity in Federated Learning. 
Although Scaffold achieves significant performance improvements, we show that this superiority is at the cost of increased security vulnerabilities. Specifically, this paper presents \name, the first backdoor attack targeting Scaffold, which turns benign clients into accomplices to amplify the attack effect. The core idea of \name is to uniquely tamper with the control variate to subtly steer benign clients' local gradient updates towards the attacker's poisoned direction, effectively turning them into unwitting accomplices, significantly enhancing the backdoor persistence. Additionally, \name leverages a GAN-enhanced poisoning strategy to enrich the attacker’s dataset, maintaining high accuracy on both benign and backdoored samples while remaining stealthy. Extensive experiments demonstrate that \name achieves superior attack durability, maintaining effectiveness for over 60 global rounds—lasting up to three times longer than existing baselines even after ceasing malicious model injections.
\end{abstract}

%

\section{Introduction}

Federated Learning (FL) enables distributed model training while preserving client data privacy. However, the effectiveness of FL models heavily depends on the distribution of training data across clients. Two scenarios typically arise: 1) IID data, where training data is uniformly distributed across clients, and 2) non-IID data, a more realistic setting where data characteristics vary significantly across clients. For IID scenarios, FedAvg \cite{mcmahan2017communication} stands out as the leading FL method, setting the standard for server-side model updates by aggregating model parameters from clients. However, its performance deteriorates in non-IID scenarios, where data heterogeneity causes update drifts from individual clients, ultimately degrading convergence \cite{li2019convergence}. 

To address this challenge, Scaffold~\cite{karimireddy2020scaffold} was introduced as a robust FL method designed to mitigate client update drift through a correction mechanism based on control variates, thereby enhancing model convergence in non-IID settings. The control variate is essentially an estimate of the difference between a client's local gradient and the global gradient, which helps align the local updates with the global objective. Scaffold reduces variance in the updates caused by data heterogeneity, making it particularly effective for scenarios where clients have diverse data distributions.

However, Scaffold Federated Learning (SFL)  not only changes the way FL models converge but also affects their robustness against adversarial manipulations. In particular, malicious clients in FL can exploit model update mechanisms to introduce backdoor behaviors, embedding hidden misbehavior into the global model~\cite{chen2017targeted}. While backdoor attacks have been extensively studied in FL~\cite{bagdasaryan2020backdoor,xie2019dba,sun2019backdoor,wang2020backdoor,dai2023chameleon}, most existing works focus on IID scenarios where attackers have full knowledge of the dataset distribution and can easily craft poisoned updates. In contrast, non-IID data distributions introduce additional constraints, making it harder for attackers to align poisoned models with the global model without significantly degrading overall performance. Although recent studies have explored backdoor attacks in non-IID FL \cite{alam2023perdoor,jeong2021abc,naseri2023badvfl,ye2024bapfl}, they have largely overlooked the unique security implications introduced by SFL. The question this paper aims to address is: \textit{if the new mechanisms of SFL (i.e., control variate for update drift correction) can bring new security threats, and unintentionally facilitate backdoor attacks in non-IID settings?} 

Our answer to the above question is affirmative. Our new insight is that \textbf{Scaffold’s reliance on control variates introduces a novel attack surface}: its correction mechanism, designed to stabilize training by aligning local updates with the global objective, can inadvertently amplify the impact of malicious updates. More critically, this mechanism allows an attacker to influence the control variate itself, effectively co-opting benign clients to ``\textit{assist in the mischief}''. Since all clients use the control variate to adjust their local gradients during updates, a tampered variate can subtly steer these honest clients’ gradients toward the attacker’s poisoned direction. This amplifies the backdoor’s reach, making Scaffold more susceptible to sophisticated attacks than standard FL methods like FedAvg, which lack such a correction mechanism.

\begin{figure}[t]
\centerline{\includegraphics[width=\linewidth]{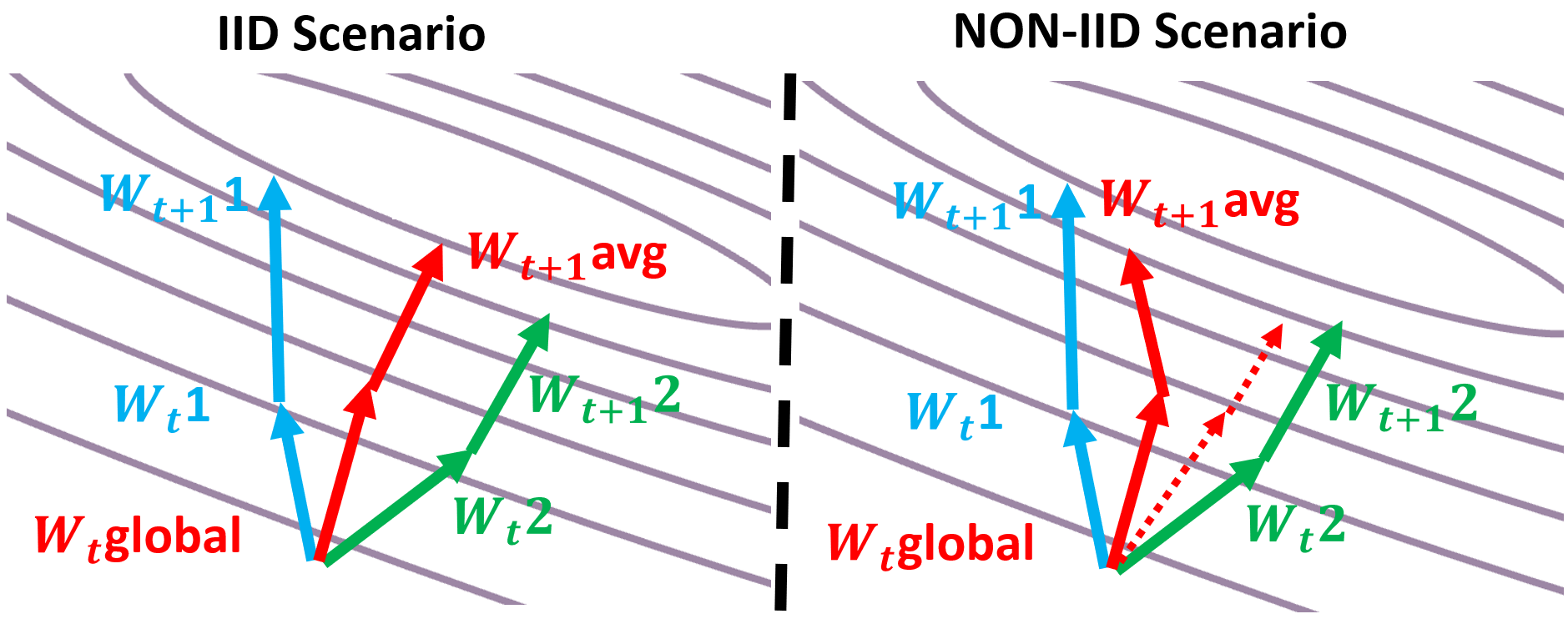 }}
\caption{Model averaging under IID and non-IID scenarios.}
\label{NON-IID2}
\vspace{-10pt}
\end{figure}

To exploit the above insight, we propose \name, a novel \ul{ba}ck\ul{d}oor attack specifically 
targeting \ul{S}caffold \ul{F}ederated \ul{L}earning, to successfully implant a backdoor function into a global model without catastrophically corrupting model performance on benign sample inference. 
Unlike prior attacks, \name leverages Scaffold’s correction dynamics to enhance both the stealth and durability of the backdoor, revealing a critical vulnerability in SFL methods. \name operates as follows:
Firstly, as the attacker only has partial knowledge of the dataset distribution in the FL system, he leverages a GAN to generate fake samples that belong to other clients to supplement the dataset, simulating a full knowledge of the dataset distribution. With the supplemented dataset for backdoor training, he gets a backdoor model achieving high accuracy in both backdoor tasks and benign tasks. Secondly, the attacker uses a distinctive feature of a category as the backdoor trigger to maintain the attack stealthiness. Thirdly, the attacker exploits the global control variate, as a reference to predict the global model's convergence direction. This optimization significantly enhances the durability of the embedded backdoor function within the global model. 

We evaluate \name on the MNIST, CIFAR-10, and CIFAR-100 datasets, demonstrating its high accuracy on both backdoor and primary tasks. Moreover, the embedded backdoor function persists in the global model for over 60 rounds and lasts 3 times longer than baseline attacks after the attacker stops injecting malicious updates. Finally, we show that \name remains highly effective when using four defense methods simultaneously. 

The main contributions are as follows:

\begin{itemize}
    \item We propose \name, the \textit{first} backdoor attack against SFL on non-IID scenarios.
    \item We enhance the backdoor durability, ensuring it persists for over 60 rounds and lasts $3\times$ longer than baselines.
    \item We conduct extensive experiments on three benchmark datasets, demonstrating high effectiveness of our attack.
\end{itemize}

\section{Background and Related Work}

\subsection{Non-IID Scenarios in FL}
In FL, non-IID refers to significant differences in data distributions among clients~\cite{kairouzs,li2021survey,mendieta2022local}. 
This discrepancy between local data distributions in non-IID scenarios can lead to inconsistencies between the local optima and the global optima. This inconsistency results in a drift in local model updates, where local models move towards their own local optima which can be far from the global optima \cite{karimireddy2020scaffold}. Consequently, averaging these local models may yield a global model far from the true global optima \cite{li2020federated,wang2020federated,wang2020fednova,karimireddy2020scaffold,lee2022preservation}, especially with numerous local epochs. As shown in Figure~\ref{NON-IID2}, while the global optima aligns with the local optima in IID scenarios, non-IID can cause the global optima to be distant from individual local optima, which is known as the \textit{client-drift} phenomenon, leading to slow and unstable convergence in the FL training process.




\begin{algorithm}
\caption{Scaffold Algorithm in Federated Learning} \label{SFL}
\textbf{Sever Input:} local datasets $D^i$, number of client $K$, number of communication rounds $R$, number of local epochs $E$ \\
\textbf{Client Input:} local control variates $c_i$, local step-size $\eta_l$ \\
\textbf{Server Updates:} \\
\hspace*{1em}$c^t \gets 0$; \\
\hspace*{1em}\textbf{for} each round r = 1, ..., $R$ \textbf{do}: \\
\hspace*{1em}\hspace*{1em} randomly selected clients $S^t \subseteq \{ 1, \ldots, K \}$ \\
\hspace*{1em}\hspace*{1em} \textbf{for} $i \in S^t $ \textbf{ in parallel do}: \\
\hspace*{1em}\hspace*{2em} send $w^t, c^t \rightarrow i$ \\
\hspace*{1em}\hspace*{2em} $\Delta w^t_i, \Delta c^t_i \gets$ \textbf{Local Update}($i, w^t, c^t$) \\
\hspace*{1em}\hspace*{1em} $w^t+1 \gets w^t - \eta \sum_{i\in S^t} \Delta w^t_i $ \\
\hspace*{1em}\hspace*{1em} $c^t+1 \gets c^t + \frac{1}{K} \Delta c $ \\
\hspace*{1em}\textbf{end for} \\ 

\textbf{Local Updates:} \\
\hspace*{1em}Local client $i$ get $w^t, c^t$ from server \\
\hspace*{1em}training model with $D_i$ get \textbf{gradient} $g_i(w_i)$ \\
\hspace*{1em}update local model: \\
\hspace*{2em} $w_i \gets w_i - \eta_l*(g_i(w_i) - c_i + c)$ \\
\hspace*{1em}update local control variates: \\
\hspace*{2em} $c^{t+1}_i \gets (i) \hspace*{1em} g_i(w_i), or \hspace*{1em} (ii) \hspace*{1em} c_i - c + \frac{1}{K*\eta_l}*(w^t - w_i)$ \\
\hspace*{1em}$\Delta w^t_i \gets w_i - w^t$ \\
\hspace*{1em}$\Delta c^t_i \gets c^{t+1}_i - c_i$ \\ 
\hspace*{1em}\textbf{return} ($\Delta w^t_i, \Delta c^t_i$)
\end{algorithm}

\noindent\textbf{Scaffold.}
Several FL algorithms have been proposed to address the above challenges, with Scaffold~\cite{karimireddy2020scaffold} being the most practical solution. It tackles the client-drift problem through the control variates (variance reduction techniques) for both the server and the clients. These control variates estimate the update direction of the global model and local client models and serve to correct local updates based on the drift, thereby mitigating the divergence between local and global optima (Alg.~\ref{SFL}). In this paper, we mainly focus on designing backdoor attacks targeting SFL.


\subsection{Backdoor Attacks against FL}
Backdoor attacks pose a significant threat to deep learning models, where malicious clients embed hidden triggers within models that cause misclassification during inference while maintaining normal performance on clean data \cite{li2022backdoor}. In FL, adversaries can deploy backdoor attacks by exploiting compromised clients to manipulate local updates, thereby generating poisoned models that corrupt the global model upon aggregation \cite{bagdasaryan2020backdoor,sun2019backdoor,wang2020backdoor,xie2019dba,zhang2022neurotoxin,li2021modelcontrastive}. 

Backdoor attacks typically involve a combination of model replacement and data poisoning. In model replacement, the adversary substitutes legitimate models with manipulated ones \cite{bagdasaryan2020backdoor}, while data poisoning involves injecting poisoned data containing the backdoor trigger into the training datasets of compromised clients. To inject these triggers, the adversary can manipulate the dataset by flipping data labels or adding a unique pixel pattern to the training samples \cite{bagdasaryan2020backdoor,wang2020backdoor,Alam}. Afterward, he strategically adjusts training parameters and scales updates to optimize the impact of the attack while evading detection by the anomaly detector deployed at the central aggregation server \cite{blanchard2017byzantine,nguyen2022flame}.

In this paper, we focus on data poisoning-based backdoor attacks. While most existing backdoor attacks target FL under IID scenarios~\cite{zhang2023a3fl,fang2023vulnerability,nguyen2023iba,xie2019dba,lyu2023poisoning}, real-world FL deployments often involve non-IID distributions, posing additional challenges for effective backdoor injection. To the best of our knowledge, no prior research has specifically explored backdoor attacks against FL with the Scaffold aggregation algorithm. We bridge this gap by investigating a novel backdoor attack targeting SFL, leveraging its unique control variate mechanism to enhance the effectiveness, stealthiness, and persistence of the attack.

\subsection{Threat Model}
\noindent\textbf{Attack scenarios.} 
We consider an attacker who aims to inject backdoors into SFL, make the final model predict the desired wrong output over a triggered input. The attacker has partial knowledge of the full dataset in the training stage. Specifically, the attacker trains a local model with a backdoor trigger function and submits poisoned local updates along with the control variate to the server for Scaffold aggregation. During inference, the attacker manipulates predictions to produce the attacker's desired outputs when inputs meet the trigger conditions.


\noindent\textbf{Attack goal.} The attack goal can be summarized as follows: 

\begin{packeditemize}
    \item \textit{Effectiveness}: The attacker must ensure that the backdoor function does not compromise the global model’s performance on primary tasks, maintaining high accuracy for both backdoor and benign predictions. 
    \item \textit{Robustness}: the backdoor should be robust against potential defenses. 
    \item \textit{Durability}: The backdoor should remain effective in the global model for as long as possible, even after the attacker stops participation in the training process, thereby maximizing the longevity of the attack.
\end{packeditemize}

\noindent\textbf{Attacker's capability.}
We assume the attacker can compromise at least one client during the training process, thereby allowing him to operate covertly within the system. Additionally, participation in FL provides the attacker with full knowledge of the model structure, facilitated by the consensus among all clients on a common learning objective. This enables the insertion of backdoor triggers, the modification of sample labels, and the manipulation of local training updates.
It is essential to note that the attacker cannot control the server or directly manipulate the aggregation procedure or the global model. They also lack access to data and models from non-compromised clients.

\begin{figure}[t]
	
    \includegraphics[width=\linewidth]{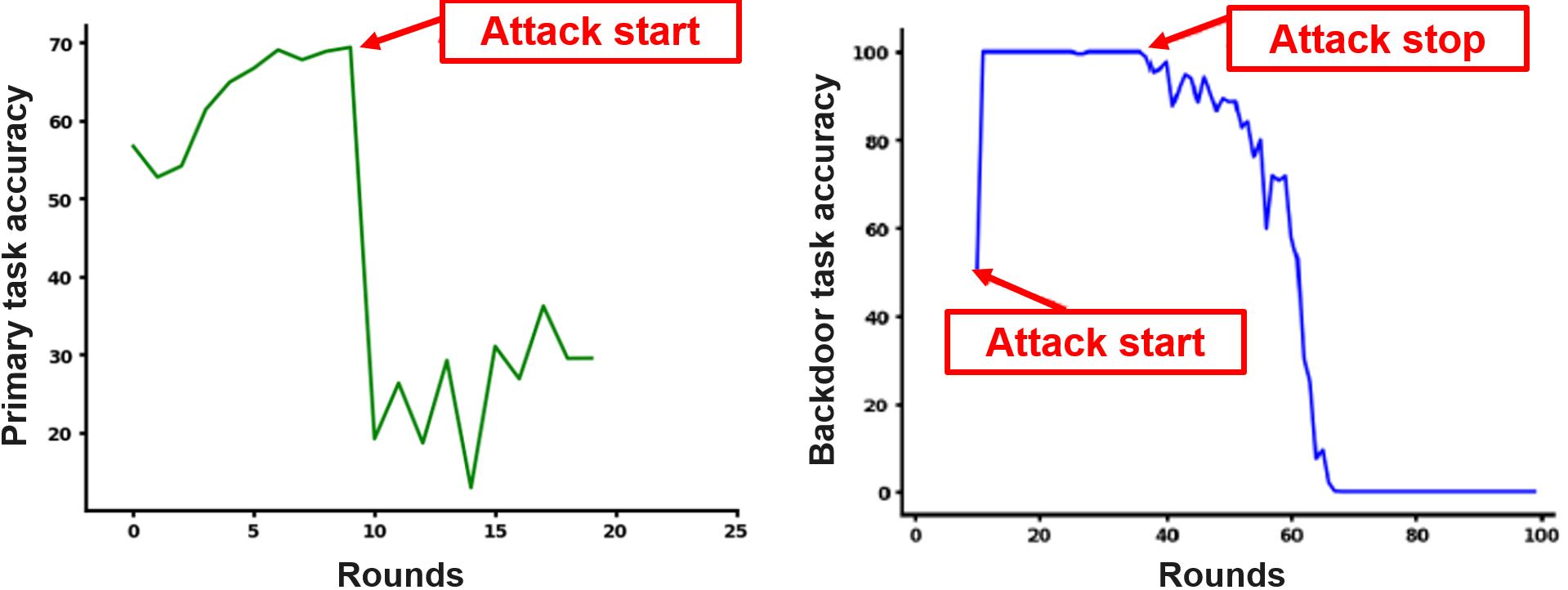}
    \caption{\textit{Left}: Primary task accuracy crushed by simple attack; \textit{Right}: Backdoor accuracy dropped after stop attacking.}
	\label{corrupt_forggeting}
 \vspace{-10pt}
\end{figure}

\section{Challenges with Backdoor Attacks in SFL} \label{sec:challenge}
Performing a backdoor attack in SFL presents the following challenges. \textbf{\one~Limited knowledge.} In non-IID scenarios, a primary challenge arises from the attacker's lack of knowledge of the dataset distribution across clients. Unlike IID scenarios, where a centralized understanding of the dataset facilitates manipulation, non-IID scenarios involve decentralized and diverse data distributions. This results in three issues: (1) Direct backdoor strategies can cause significant performance degradation on benign samples, leading to the rejection of the global model; (2) The variability in data distributions increases the difference between local and global models, making malicious models more detectable;  (3) Averaging poisoned models with the global model degrades its performance on the primary task, as shown in Figure \ref{corrupt_forggeting}, where accuracy drops significantly when a poisoned model is aggregated. \textbf{\two~Control variate.} In SFL, control variate (denoted as $c_i$) is used to correct the client drift and align local models with the global model.  If attackers strictly follow protocols and use the $c_i$) to correct their malicious models during the triggering planting process, the effectiveness of the attack can be reduced. Conversely, if an attacker chooses to manipulate the $c_i$ inappropriately, introducing a malicious $c$ to the server, it could lead to a potential corruption of the global model. \textbf{\three~Backdoor catastrophic forgetting.} Catastrophic forgetting~\cite{kirkpatrick2017catastrophic} occurs when neural networks forget previously learned tasks upon learning new ones. 
This can cause backdoors to lose effectiveness over time~\cite{wang2020backdoor}. If attackers stop uploading malicious updates, the backdoor function may eventually be erased by benign updates. As shown in Figure \ref{corrupt_forggeting}, the accuracy of backdoor tasks declines sharply over time, with the backdoor function vanishing around round 65. Although various methods \cite{zhang2022neurotoxin,alam2023perdoor,dai2023chameleon,wen2022anticipate} have been proposed to address this issue, none have proven effective in SFL.



\begin{algorithm}
\caption{\name} \label{BadSFLAlgo}
\textbf{Required:} local datasets $D^i$, global model $w_g$, global control variate $c$, number of local epochs $E$, local learning rate $\eta_l$, Generator $G$, Discriminator $D$ \\ 
\hspace*{1em}Update local model with $w_p \gets w_g$ \\
\hspace*{1em}Initialize Discriminator $D \gets w_g$ \\
\hspace*{1em}\textbf{do:} \\
\hspace*{1em}\hspace*{1em}Run G for generating fake samples \\
\hspace*{1em}\hspace*{1em}Evaluate fake sample on D \\
\hspace*{1em}\hspace*{1em}Update G using D \\
\hspace*{1em}\textbf{until} G converges to generate target samples \\ 
\hspace*{1em}G generates samples into dataset $D_f$ \\
\hspace*{1em}$D_c \gets D_f + D^i$ \\
\hspace*{1em}Select backdoor samples from $D_c$ and assign them wrong label as $D_b$\\
\hspace*{1em}$D_p \gets D_c + D_b$ \\
\hspace*{1em}\textbf{for} each epoch e = 1, ..., $E$ \textbf{do}: \\ 
\hspace*{1em}\hspace*{1em} $w_p = \mathop{argmin}_{w_p} [L(D_p, w_p) + L(D_p, P_j(w_p, c))].$ \\
\hspace*{1em}\textbf{end for} \\
\hspace*{1em}$\Delta w_p = w_p - w_g$ \\
\hspace*{1em}$\Delta c_p = \frac{1}{K*\eta_l}*(w_g - w_p) - c$ \\
\textbf{return} ($\Delta w_p,     \Delta c_p$) 
\end{algorithm}

\section{\name}

\noindent\textbf{Overview.} To overcome the challenges, we propose \name, as detailed in Algorithm \ref{BadSFLAlgo}.
\name mainly consists of 4 steps: \textit{Step 1: Initialization.} The attacker initiates the attack by downloading the global model $w_g$ and controlling the variate $c$ from the server. Subsequently, the attacker updates the local model $w_p$ and discriminator $D$ using the downloaded global model $w_g$. \textit{Step 2: GAN-based Training for Data Supplementation.} The attacker performs GAN training on generator $G$ and discriminator $D$. The training terminates upon the convergence of the generator, signifying its ability to generate realistic fake samples in class $C$ that do not belong to $D^i$ but rather originate from other clients' datasets. Then the generator $G$ is utilized to generate a number of samples in class $c$ forming in dataset $D_f$. This dataset $D_f$ is then merged with the attacker's original dataset $D^i$ to create a new dataset $D_c$. \textit{Step 3: Backdoor Sample Selection and Trigger Injection.} With dataset $D_c$, the attacker selects specific samples with a characteristic feature to serve as backdoor samples. These samples are then relabeled to a target class $x$ as the backdoor target class that is different from their original labels. The attacker organizes these manipulated samples into a separate dataset $D_b$ and merges it with $D_c$ to finalize the dataset $D_p$ for backdoor training. 
\textit{Step 4: Backdoor Model Training and Optimization.} The attacker proceed to trains the local model $w_p$ based on the dataset $D_p$. During the training process, the attacker follows the equation \ref{bda3} to optimize the backdoor objective. Upon convergence, the backdoor model update $\Delta w_p$ and the corresponding control variate $\Delta c_p$ are obtained and can be uploaded to the server.

\subsection{GAN-based Dataset Supplementation} 

In non-IID data scenarios, directly injecting backdoor samples into dataset $D^i$ for training often leads to a more biased model, deviating significantly from global optima \cite{psychogyios2023gandriven,lai2022ganbased}. To mitigate this, inspired by Zhang et al. \cite{zhang2019gan}, the attacker can employ GAN to generate synthetic samples that resemble the data held by other clients. This involves training a generator $G$ with local non-IID data to bridge the gap between datasets. The GAN architecture basically consists of a generator $G$ and a discriminator $D$. In our case, the generator $G$ comprises a series of `deconvolution' layers that progressively transform random noise into a sample, while the discriminator $D$ closely resembles the global model, except for its output layer, which distinguishes between fake and real samples. The attacker iteratively trains the generator $G$ locally with the constraint of discriminator $D$ until it converges to generate realistic fake samples that do not belong to the attacker. Concurrently, as the SFL procedure progresses, the global model tends to converge. 
During each server-client communication round, the attacker updates the discriminator $D$ using the latest global model $w_g$ downloaded from the server and performs new-round optimization training on the generator $G$, guiding it to generate more authentic fake samples that closely resemble data from other clients. These high-quality synthetic samples are then integrated into the attacker's original non-IID dataset, effectively supplementing it with additional data classes. 

\begin{figure}[t]
\centerline{\includegraphics[width=\linewidth]{ 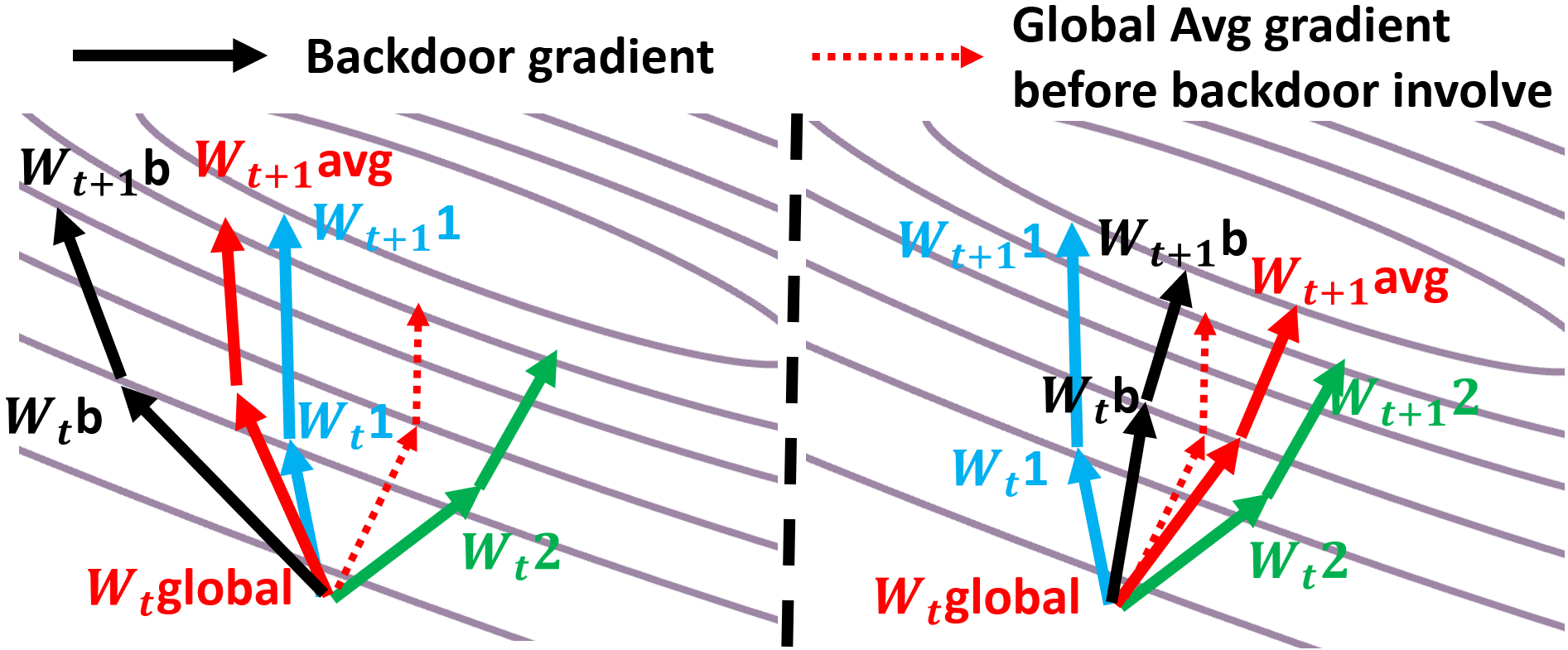}}
\vspace{-10pt}
\caption{\textit{Left}: global optima under non-IID scenario. \textit{Right}: global optima with data supplementation techniques.}
\label{datasup}
\vspace{-10pt}
\end{figure}

The attacker synchronously updates the discriminator $D$ during each local training round using the new global model $w_g$ downloaded from the server, followed by GAN training to optimize the generator $G$ for improved performance. The output of this process is then merged into the attacker's non-IID dataset for further backdoor training. As the generated samples closely resemble those from other client datasets, the attacker local optima trained by the attacker can converge closer to the global optima than other clients. Figure \ref{datasup} demonstrates the difference between the aggregated global optima with and without data supplementation techniques.

\begin{table*}[t]
\centering
\caption{Dataset, model structure, and hyperparameter description.}
\vspace{-10pt}
\begin{tabular}{|c|c|c|c|c|c|c|ccc|c|}
\hline
\multirow{2}{*}{Dataset} &\multirow{2}{*}{Instances}&\multirow{2}{*}{Features}& \multirow{2}{*}{Model} & \multirow{2}{*}{Benign $l_r$} & \multirow{2}{*}{E} & \multirow{2}{*}{Poison $l_r$} & \multicolumn{3}{c|}{Poison ratio} & \multirow{2}{*}{Batch Size}   \\ \cline{8-10}
&&&&&&&FL&PT&FB&
   \\ \hline
CIFAR-10 &60000& 1024& ResNet-18& 0.001& 10 & 0.05 &0.1 &0.0125& 0.01& 128
   \\ 
CIFAR-100&60000 & 1024 &ResNet-50& 0.0001 & 100 & 0.0001 & 0.1 &0.0125& 0.01 & 128
  \\ 
MNIST &70000& 784 & ConvNet &0.01& 2 & 0.001 & 0.1&0.0125&- &128
    \\ \hline
\end{tabular}
\label{tab:config}
\vspace{-10pt}
\end{table*}

\subsection{Trigger Selection and Injection}

With the prepared dataset $D_p$, containing both original and synthetic data, the attacker proceeds to inject a backdoor into the model. \name leverages three techniques to inject backdoors: (1) Label-flipping \cite{li2021label}, in which the ground truth labels of a whole class $a$ in $D_p$ are directly altered to another label $b$. For instance, all the `dog' labels are altered to `cat' in CIFAR-10. (2) Pattern trigger \cite{uprety2021pattern}, which involves poisoning samples with a specific trigger pattern, i.e., a small mosaic cube added in the images to activate the backdoor behavior. The attacker injects these poisoned images into the $D_p$ along with a target label, establishing a correlation between the trigger pattern and the desired misclassification. (3) A stealthier backdoor method, known as feature-based backdoor \cite{mayerhofer2022feature}, involves selecting a distinctive feature within an image class as the backdoor trigger. This approach eliminates the need to directly manipulate the images, thereby increasing the difficulty of detection. For instance, all the green cars in the `car' class in CIFAR-10 are designed as the backdoor trigger. During the inference stage, the compromised model outputs the attacker's target label only when the input is an image containing a green car. The selection of a unique feature within a class makes this trigger difficult to detect as it appears as a natural variation within the data.

\subsection{Backdoor Training with Control Variate}

\begin{figure}[t]
\centerline{\includegraphics[width=0.45\linewidth]{ 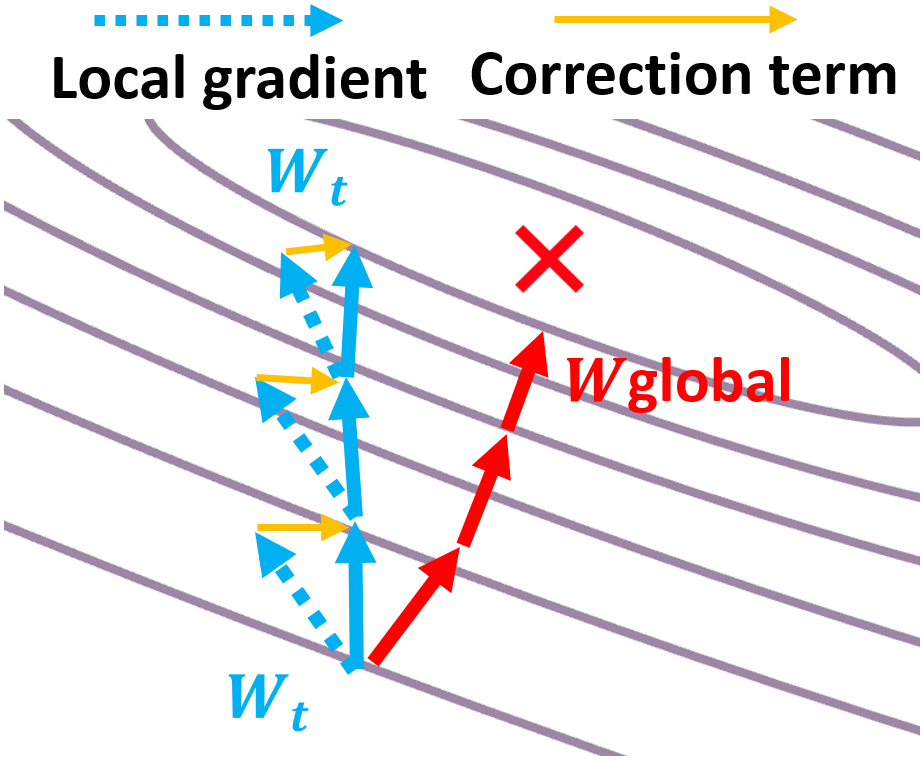}}
\caption{Scaffold correction term on a single client.}
\label{correctionterm}
\end{figure}

As discussed in section~\ref{sec:challenge}, the global control variate $c$ is utilized in SFL to correct the client drift. Specifically, the correction value $c-c_i$ adjusts the local model point towards the global model, as shown in algorithm \ref{SFL}. During the local model training process, this correction term effectively `drag' the drifting local model closer to the global model, facilitating convergence towards the global optima, as depicted in Figure \ref{correctionterm}. In the server aggregation round, the global control variate $c$ is computed by averaging the drift values of all local models, which represents the convergence direction of the global model.

From the attacker's perspective, allowing the control variate to correct the poisoned model according to SFL rules can reduce the effectiveness of the backdoor attack, as discussed in Section~\ref{sec:challenge}. However, the attacker still needs to submit a control variate $c_p$ to report the drift of the backdoor model. The key idea is to train a backdoor model that is closer to the global model compared to other local models trained on non-IID data. Since the global control variate $c$ is known to participating clients, it can be used as a reference for the global model's convergence direction, helping to align the poisoned model more closely with the global optimum. This constraint, derived from $c$, functions similarly to using the future global model for optimization, as suggested by Wen et al.\cite{wen2022anticipate}. This constraint can be integrated into the loss function to enhance the backdoor's effectiveness and persistence in the global model.

Initially, the attacker performs backdoor training and optimizes their backdoor objective as in the Equation \ref{bda1} \cite{bhagoji2019opt}:
\begin{equation}
w^*_p = \mathop{argmin}_{w_p}L(D_p, w_p).
\label{bda1}
\end{equation}
where $L$ is the loss function of the backdoor task, $w_p$ is the attacker model weights.

In our \name attack, we modify the standard backdoor objective function by adding a term to ensure that the backdoor updates sent to the server persist in the backdoor function in the global model for more future training rounds. We can simulate an aggregation round and apply the control variate $c$ to obtain a predicted global model for one future round. Here is the modified objective function (Equation \ref{bda2}):
\begin{equation}
P_j(w_p, c) = \frac{w_p + w_g * (n-1)}{n} - \eta _l * c * j
\label{bda2}
\end{equation}
To summarize, we formalize our attack objective as below:
\begin{equation}
w^*_p = \mathop{argmin}_{w_p} [L(D_p, w_p) + L(D_p, P_j(w_p, c))].
\label{bda3}
\end{equation}
where $j$ represents the number of future rounds that $w_p$ anticipates. By optimizing the backdoor model to be closer to the global model, the attacker simultaneously optimizes the control variate $c_p$ to align it with the expected drift value. This ensures that the attacker's actions conform to the SFL protocol (Algorithm \ref{SFL}).

\section{Evaluation}
\subsection{Setup}
We evaluate \name on a server running Ubuntu 18.04 with an NVIDIA GeForce RTX 2080 Ti GPU. 

\noindent\textbf{Datasets, models, and hyperparameters.}
We consider the datasets that are commonly used in previous works~\cite{lyu2023poisoning}, i.e.,  MNIST~\cite{6296535}, CIFAR-10~\cite{krizhevsky2009learning} and CIFAR-100~\cite{krizhevsky2009learning}.  Detailed configurations are summarized in Table~\ref{tab:config}.

\noindent\textbf{GAN models.}
In attacking SFL with CIFAR-10 and CIFAR-100, the discriminator mimics ResNet18 and ResNet50, respectively, differing only in the output layer. The generator consists of five deconvolutional layers to generate synthetic images for supplementing \( D_p \) in the backdoor attack. For MNIST, the discriminator follows LeNet-5, while the generator uses three deconvolutional layers to produce synthetic images resembling those of other clients.

\noindent\textbf{Attack bsaselines.}
We compare \name with 4 backdoor attacks: (1) \textit{Black-box Attack} \cite{bagdasaryan2020backdoor}. This attack directly poisoned the training dataset through techniques such as label-flipping, pattern trigger, and feature-based trigger. Subsequently, it conducts backdoor training to minimize the classification loss on the dataset $D_p$. (2) \textit{Neurotoxin} \cite{zhang2022neurotoxin}. It incorporates a strategic approach to ensure the durability of the backdoor function within the global model. Specifically, it aims to prevent the malicious model updates from pointing toward coordinates that are frequently updated by benign clients, thereby mitigating the risk of the backdoor being erased. We illustrate it in Figure~\ref{neurotoxinfig} in the Appendix. (3) Irreversible Backdoor Attack (IBA)~\cite{nguyen2023iba}. IBA gradually implants a stealthy and durable backdoor into the global model by optimizing trigger imperceptibility and selectively poisoning parameters less likely to be updated. 
(4) 3DFed~\cite{li20233dfed}. 3DFed is a multi-layered backdoor attack framework that combines constrained loss training, noise masking, and a decoy model to evade detection in a black-box FL setting.

\begin{figure}[t]
    \centering
    \begin{subfigure}{0.48\linewidth}
        \centering
        \includegraphics[width=\linewidth]{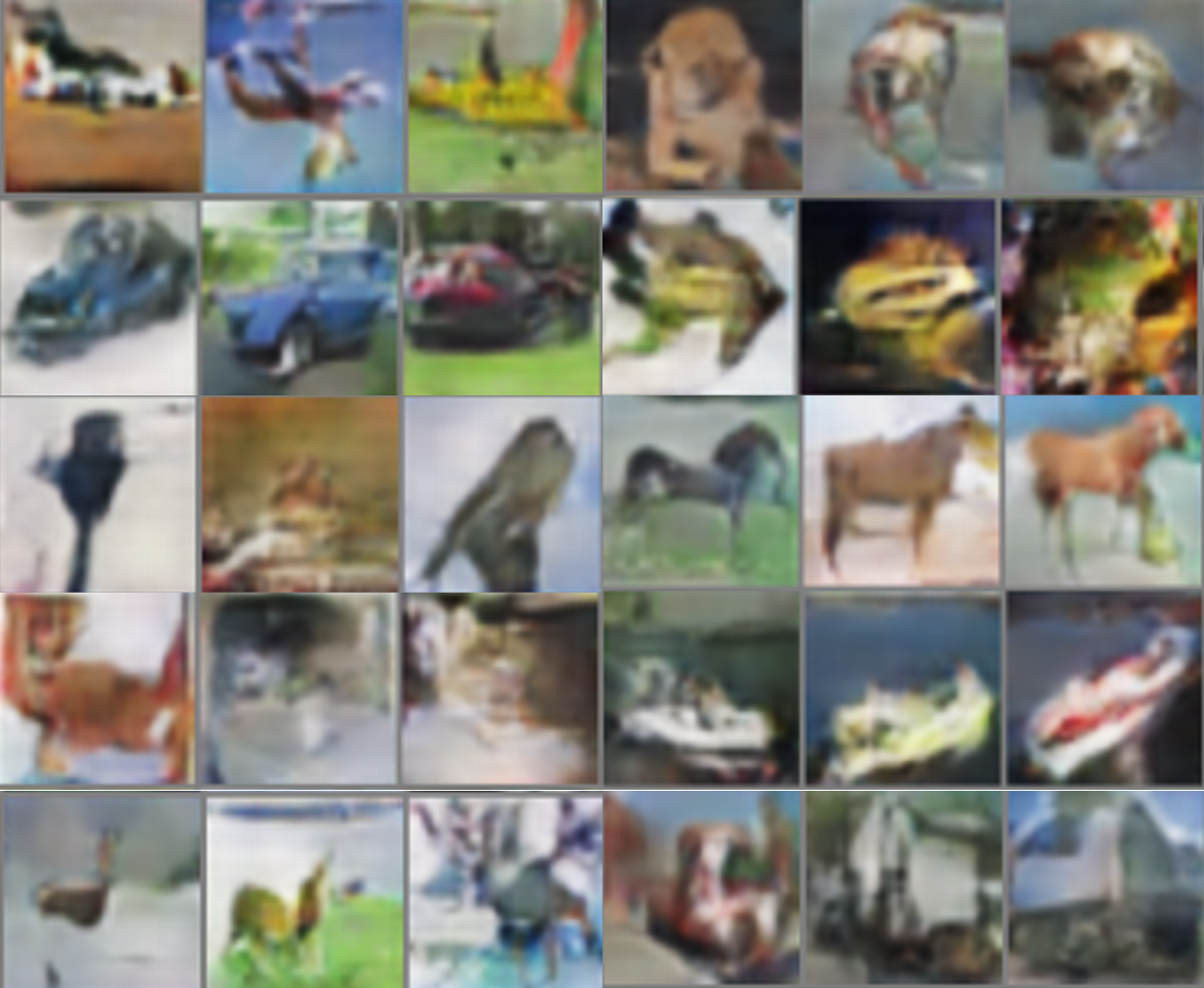}
        \caption{\centering Generated samples by GAN.}
        \label{cifarfake}
    \end{subfigure}
    \hfill
    \begin{subfigure}{0.48\linewidth}
        \centering
        \includegraphics[width=\linewidth]{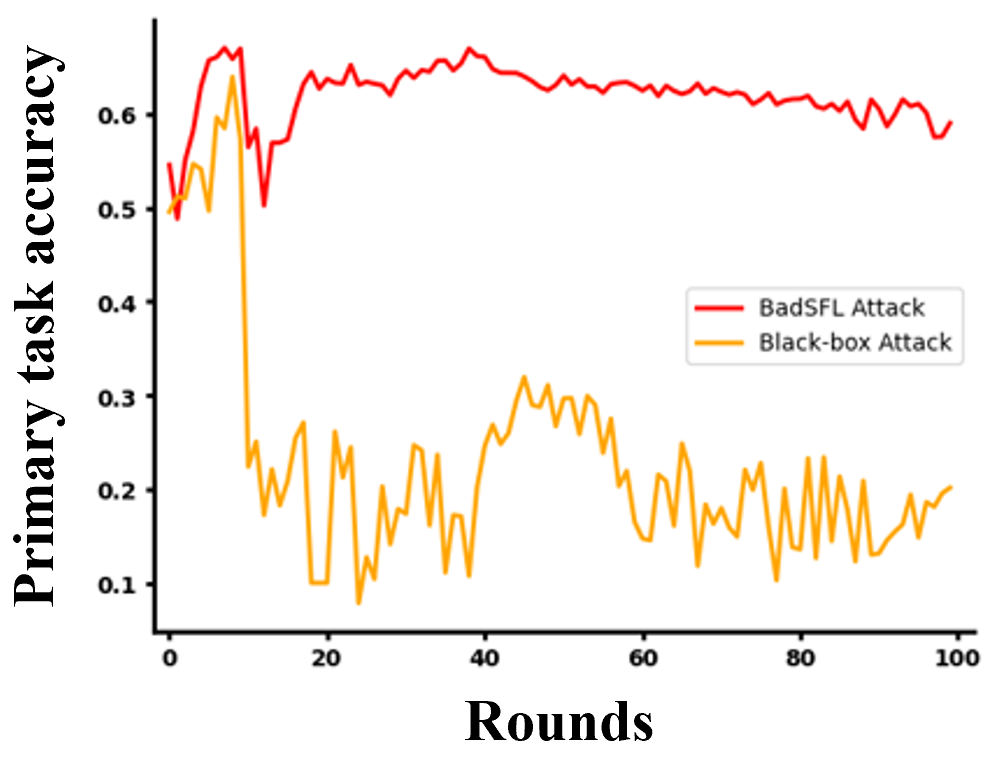}
        \caption{PTA on CIFAR-10.}
        \label{cifarexp1}
    \end{subfigure}


    \begin{subfigure}{0.48\linewidth}
        \centering
        \includegraphics[width=\linewidth]{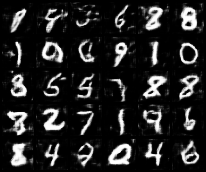}
        \caption{\centering Generated samples by GAN.}
        \label{mnistfake}
    \end{subfigure}
    \hfill
    \begin{subfigure}{0.48\linewidth}
        \centering
        \includegraphics[width=\linewidth]{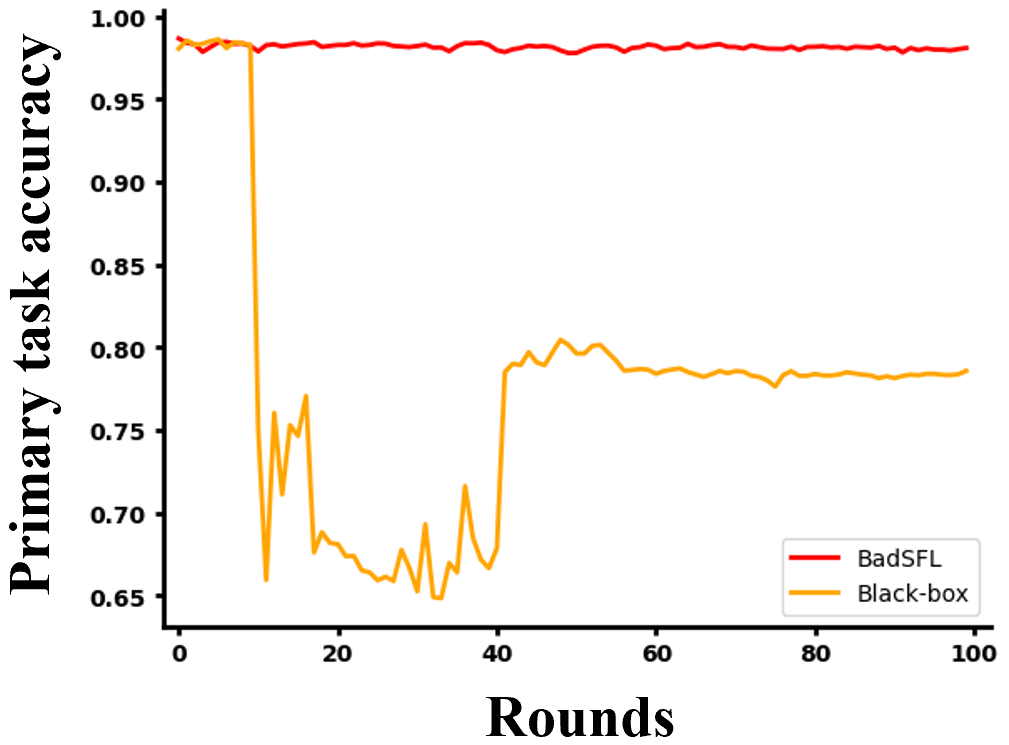}
        \caption{PTA on MNIST.}
        \label{mnistexp1}
    \end{subfigure}
    \caption{Dataset supplementation on CIFAR-10 and MNIST.}
    \vspace{-10pt}
\end{figure}

\begin{figure*}[t]
    \begin{subfigure}{0.23\linewidth}
        \includegraphics[width=\linewidth]{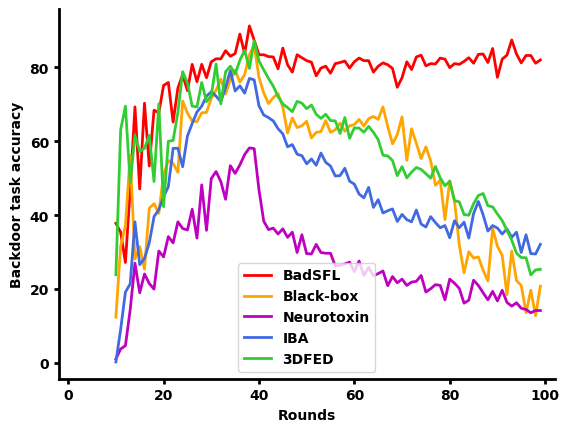}
        \caption{Label Flipping (CIFAR-10)}
        \label{a}
    \end{subfigure}
    \hfill
    \begin{subfigure}{0.23\linewidth}
        \includegraphics[width=\linewidth]{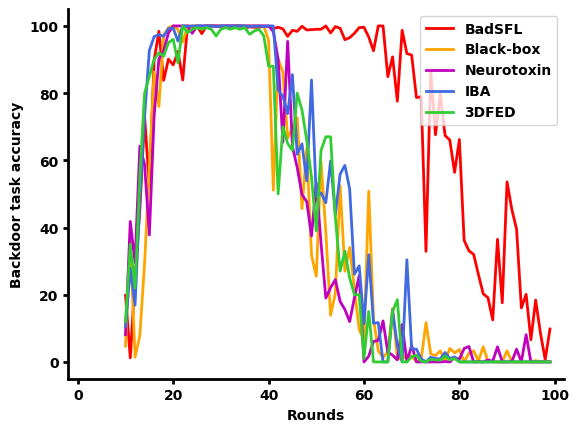}
        \caption{Pattern Trigger (CIFAR-10)}
        \label{b}
    \end{subfigure}
    \hfill
    \begin{subfigure}{0.23\linewidth}
        \includegraphics[width=\linewidth]{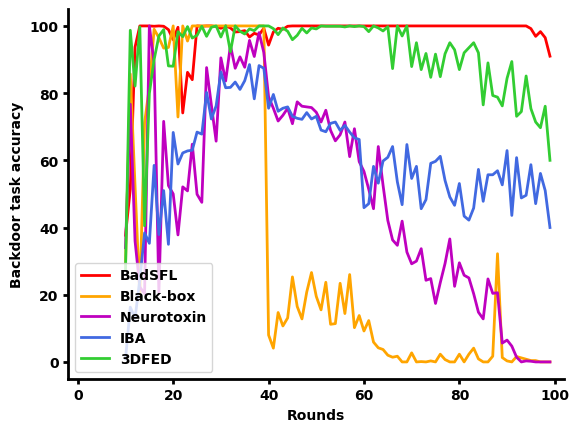}
        \caption{Plane in Sunset (CIFAR-10)}
        \label{c}
    \end{subfigure}
    \hfill
    \begin{subfigure}{0.23\linewidth}
    \includegraphics[width=\linewidth]{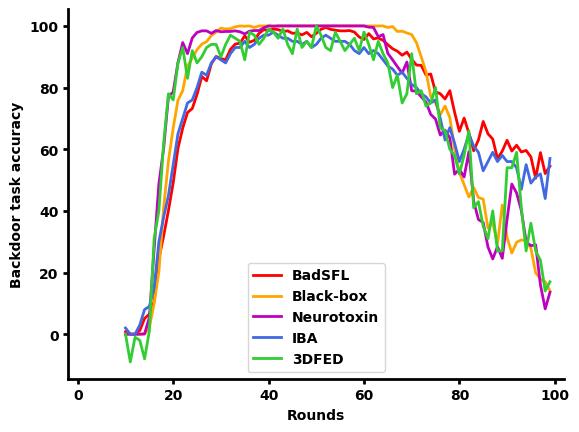}
    \caption{Label Flipping (CIFAR-100)}
    \label{d}
    \end{subfigure}

    \begin{subfigure}{0.23\linewidth}
        \includegraphics[width=\linewidth]{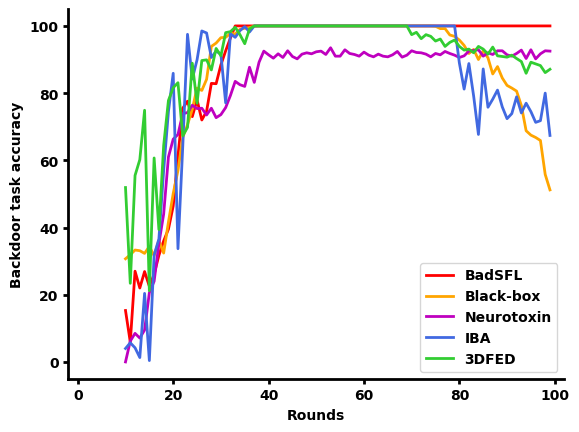}
        \caption{Plane in Sunset (CIFAR-100)}
        \label{e}
    \end{subfigure}
    \hfill
    \begin{subfigure}{0.23\linewidth}
    \includegraphics[width=\linewidth]{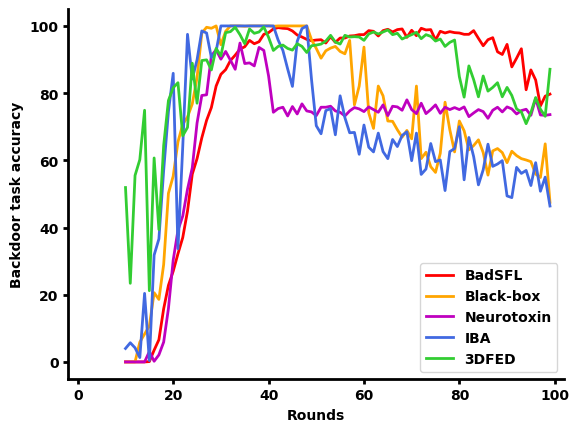}
    \caption{Yellow Truck (CIFAR-100)}
    \label{f}
    \end{subfigure}
        \hfill
    \begin{subfigure}{0.23\linewidth}
        \includegraphics[width=\linewidth]{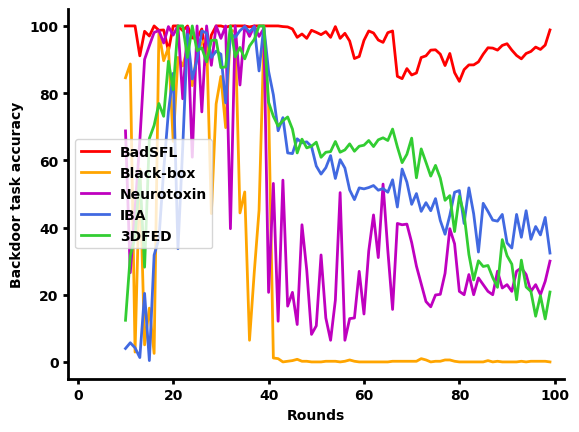}
        \caption{Label Flipping (MNIST)}
        \label{g}
    \end{subfigure}
    \hfill
    \begin{subfigure}{0.23\linewidth}
        \includegraphics[width=\linewidth]{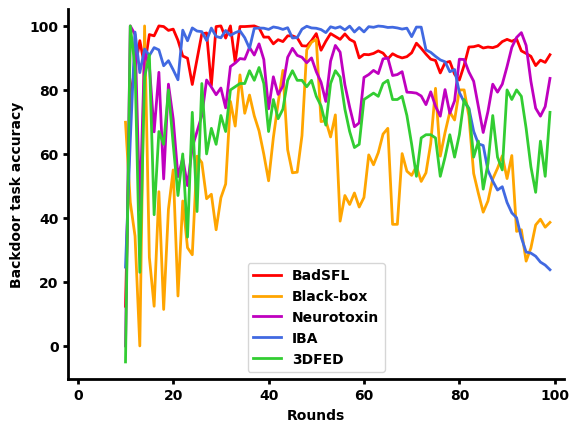}
        \caption{Pattern Trigger (MNIST)}
        \label{h}
    \end{subfigure}
    


    
    \caption{Attack comparisons with baselines.}
    \label{mnistallexp}
    \vspace{-10pt}
\end{figure*}

\begin{figure}[t]
    \begin{subfigure}{0.45\linewidth}
        \centering
        \includegraphics[width=\linewidth]{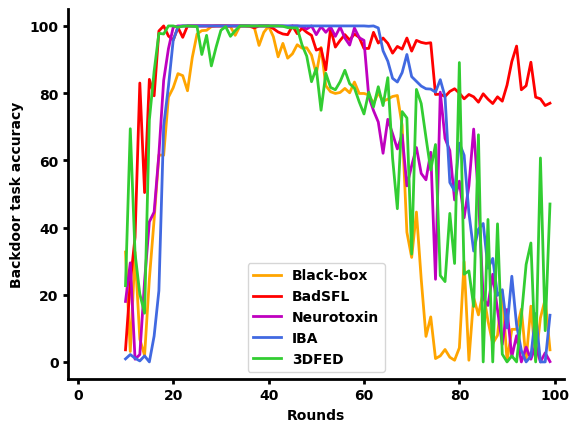}
        \caption{Green Car}
        \label{expgreencar}
    \end{subfigure}
    \hspace{5pt}
    \begin{subfigure}{0.45\linewidth}
        \centering
        \includegraphics[width=\linewidth]{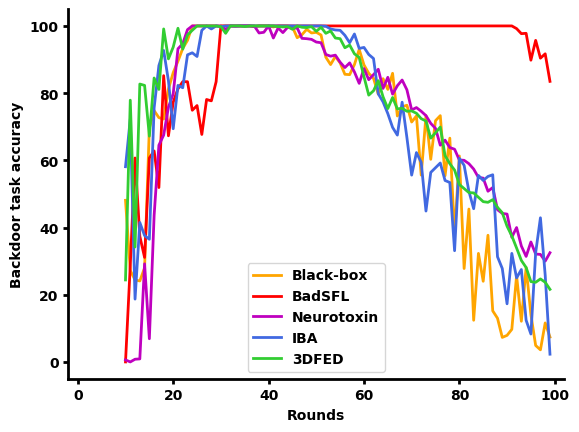}
        \caption{Race Car}
        \label{expredcar}
    \end{subfigure}
    \medskip

    \begin{subfigure}{0.45\linewidth}
        \centering
        \includegraphics[width=\linewidth]{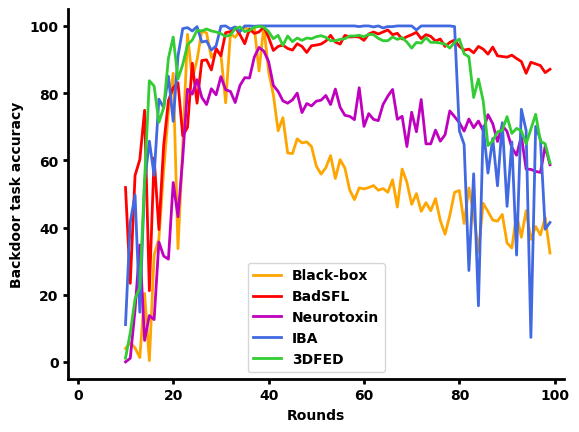}
        \caption{White Horse}
        \label{exphorse}
    \end{subfigure}
    \hspace{5pt}
    \begin{subfigure}{0.45\linewidth}
        \centering
        \includegraphics[width=\linewidth]{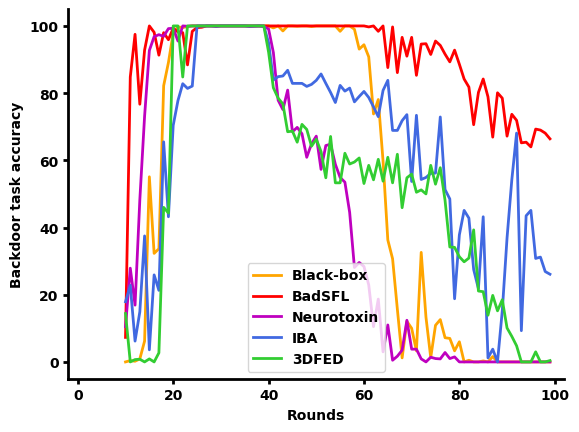}
        \caption{Red Ship}
        \label{expship}
    \end{subfigure}
    \hfill
    \vspace{-8pt}
    \caption{Feature-based backdoor attacks on CIFAR-10.}
    \vspace{-10pt}
    \label{cifarappendexp}
\end{figure}

\noindent\textbf{Attack settings.}
We run 100 communication rounds in SFL for CIFAR-10, CIFAR-100, and MNIST. In each round, 20 clients participate, with 50\% randomly of them selected for training. To ensure a non-IID data distribution, the training dataset is split into 200 label-sorted groups, which are randomly assigned to all the participating clients.

For \name evaluations on CIFAR-10 and CIFAR-100 datasets, the attacker strategically joins the training process from the 10th round, exiting after the 40th round. Initially, the attacker conducts GAN training to perform data supplementation, employing 10 local epochs, a fixed learning rate of 0.001, and Adam Optimizer with parameters (0.5, 0.999). Subsequently, backdoor training and optimization are carried out with 10 local epochs, a future round $j$ set to 10, a fixed learning rate of 0.05, and an SGD optimizer with a momentum of 0.9 and a weight decay of 0.005.

To obtain $D_p$, we employ three types of backdoor trigger injections and evaluate the performance of \name based on BTA in each communication round, contrasting it with baseline attacks: (1) Label Flipping. All `dog' samples in $D_c$ are relabeled as `bird'; (2) Pattern Trigger. A small triangle pattern is added to the right bottom of the image, with all the modified samples labeled as `cat'; (3) Feature-based trigger. Characteristic features  (e.g., car stripe, green car, race car, sunset plane, white horse, red ship, yellow truck)  are chosen as backdoor triggers in Dataset $D_c$, with corresponding samples labeled as `bird'.

For MNIST, the attacker engages in the training process from the 10th round and quits after the 40th round. Also, GAN training is employed for data supplementation, comprising 2 local epochs, a fixed learning rate of 0.001, and Adam Optimizer with parameters (0.5, 0.999). Subsequently, backdoor training and optimization are performed over 2 local epochs, considering future rounds $j$ as 10, with a fixed learning rate of 0.1 and SGD optimizer.





To obtain $D_p$, only two types of backdoor trigger injections are used, as MNIST lacks sufficient features for feature-based backdoor attacks. The performance of \name is evaluated based on the BTA in each communication round, compared to two baselines: (1) Label flipping. All samples labeled as `5' in $D_c$ are relabeled as `2'; (2) Pattern trigger. A small triangle pattern is added to the right bottom of the image, and the labels of all modified samples are changed to `2'.

\noindent\textbf{Evaluation metric.}
We adopt two metrics commonly utilized in FL. (1) Primary Task Accuracy (PTA), which reflects the classification accuracy on clean samples. (2) Backdoor Task Accuracy (BTA), which is the attack success rate, measured by the poisoned model’s accuracy on poisoned samples.


\subsection{Results of GAN}
Figure \ref{cifarfake} and Figure \ref{mnistfake} illustrate the datasets $D_f$ generated by the Generator $G$ for CIFAR-10 and MNIST, respectively. In both cases, $D_f$ encompasses nearly all classes of each dataset, demonstrating the successful acquisition of full dataset distribution knowledge within the SFL framework under non-IID scenarios. Utilizing the combined dataset $D_c$, the attacker executes the backdoor attack while maintaining the model's accuracy on the primary task. For CIFAR-10, as shown in Figure \ref{cifarexp1}, the primary task accuracy remains around 55\% with data supplementation, compared to a drop below 25\% in the baseline attack. Similarly, for MNIST, Figure \ref{mnistexp1} reveals that using $D_c$ instead of $D_i$, the primary task accuracy stays above 90\%, while the baseline attack results in a decline to under 75\%.

\subsection{Effectiveness of \name}

From Figures \ref{a} to \ref{f}, we present attack comparisons with baselines evaluated on the CIFAR-10 and CIFAR-100 datasets. 
It is evident that \name outperforms the baseline attacks in terms of both effectiveness and durability. To be specific, \name achieves above 80\% backdoor task accuracy across all types of backdoor attacks within the first 10 rounds, while the attacker remains active in the training process for backdoor training and malicious updates to the server. Meanwhile, \name keeps the primary task accuracy at 60\% (Figure \ref{cifarexp1}). Furthermore, even after the attacker exits the training process at the 40th round, the benign clients continue submitting normal updates in subsequent rounds, which could potentially affect the poisoned updates from the attacker in previous attacking rounds thus erasing the backdoor function. Despite this, \name ensures a resilient backdoor function with accuracy exceeding 90\% over the entire 100 SFL rounds, which is 3 times longer than the lifespan achieved by the two baseline attacks, where backdoor task accuracy drops below 50\% after the 60th round. 
Horizontally comparing the effects of different types of backdoor trigger injections (Figures~\ref{a}, \ref{b} and \ref{c}), it is observed the feature-based trigger performs the best among them, benefiting from its stealthiness without directly manipulating the images, thereby making its updates less likely to conflict with those from benign updates. We also provide more results available in Figure~\ref{cifarappendexp}.


\begin{figure}[t]
    \centering
    \begin{subfigure}{0.49\linewidth}
        \centering
        \includegraphics[width=\linewidth]{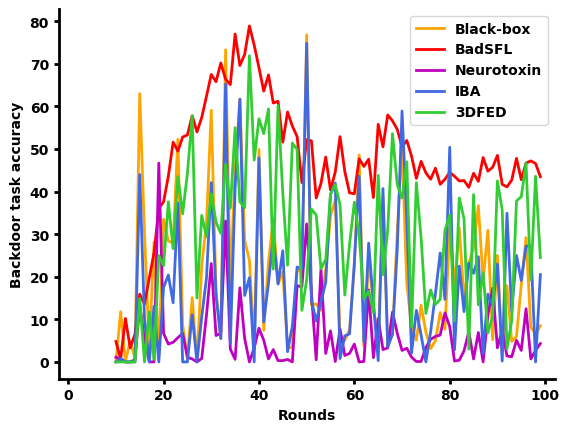}
        \caption{Label Flipping, CIFAR-10, differential privacy + model pruning.}
        \label{defense_mnistflip}
    \end{subfigure}
    \hfill
    \begin{subfigure}{0.49\linewidth}
        \centering
        \includegraphics[width=\linewidth]{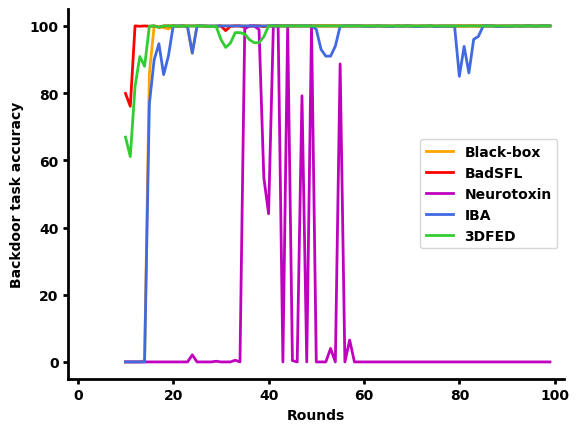}
        \caption{Pattern Trigger, CIFAR-10, differential privacy + model pruning.}
        \label{defense_patterntrigger}
    \end{subfigure}

   \medskip

    \begin{subfigure}{0.49\linewidth}
        \centering
        \includegraphics[width=\linewidth]{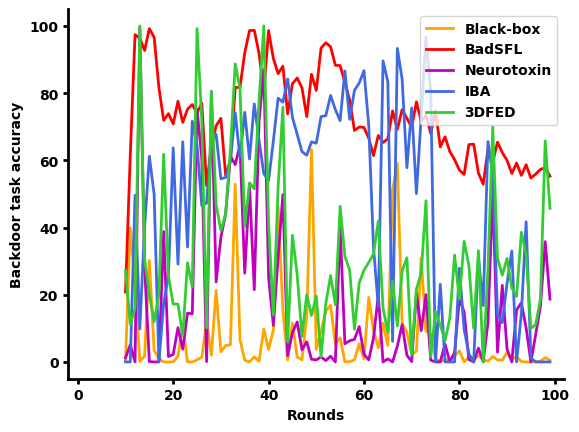}
        \caption{ Feature-based Trigger, CIFAR-10, differential privacy + model pruning.}
        \label{defense_mnisttriggr}
    \end{subfigure}
    \hfill
    \begin{subfigure}{0.49\linewidth}
        \centering
        \includegraphics[width=\linewidth]{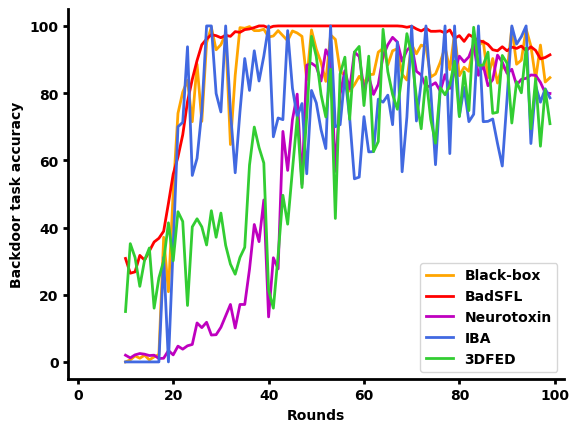}
        \caption{Feature-based Trigger, CIFAR-100, differential privacy + model pruning.}
        \label{defense_mnisttrigger_100}
    \end{subfigure}

    \begin{subfigure}{0.49\linewidth}
    \centering
    \includegraphics[width=\linewidth]{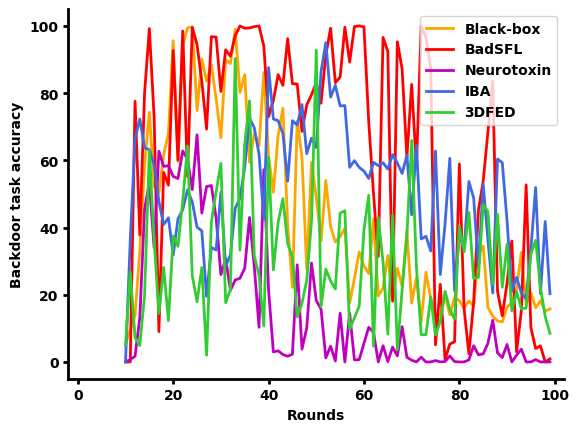}
    \caption{ Feature-based Trigger, CIFAR-100, FLAME.}
    \label{defense_cifar10sunsetplane_100}
    \end{subfigure}
    \hfill
    \begin{subfigure}{0.49\linewidth}
        \centering
        \includegraphics[width=\linewidth]{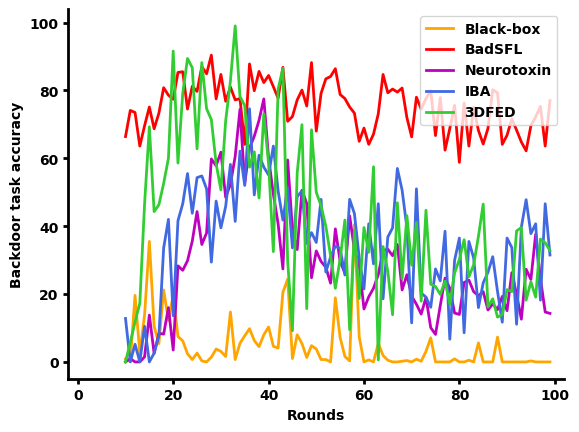}
        \caption{Feature-based Trigger, CIFAR-100, SparseFED.}
        \label{defense_mnisttrigger_100}
    \end{subfigure}

    
    \hfill
    \vspace{-10pt}
    \caption{Defense against backdoor attacks.}
    \vspace{-20pt}
    \label{defense}

\end{figure}

Figures \ref{g} and \ref{h} showcase the results obtained on the MNIST dataset. Similarly, \name outperforms the other baseline attacks, achieving both backdoor task accuracy and primary task accuracy above 85\%. After the malicious update injection stops at round 40, in the Label flipping attack, the backdoor task accuracy in two baseline attacks catastrophically drops below 40\% within 10 rounds whereas \name maintains a 5 times longer-lasting backdoor function in the global model in future rounds. In the pattern trigger attack, \name also injects a more effective backdoor function into the global model with 10\% higher accuracy compared to the baselines.

\begin{figure}[t]
\centerline{\includegraphics[width=0.75\linewidth]{ 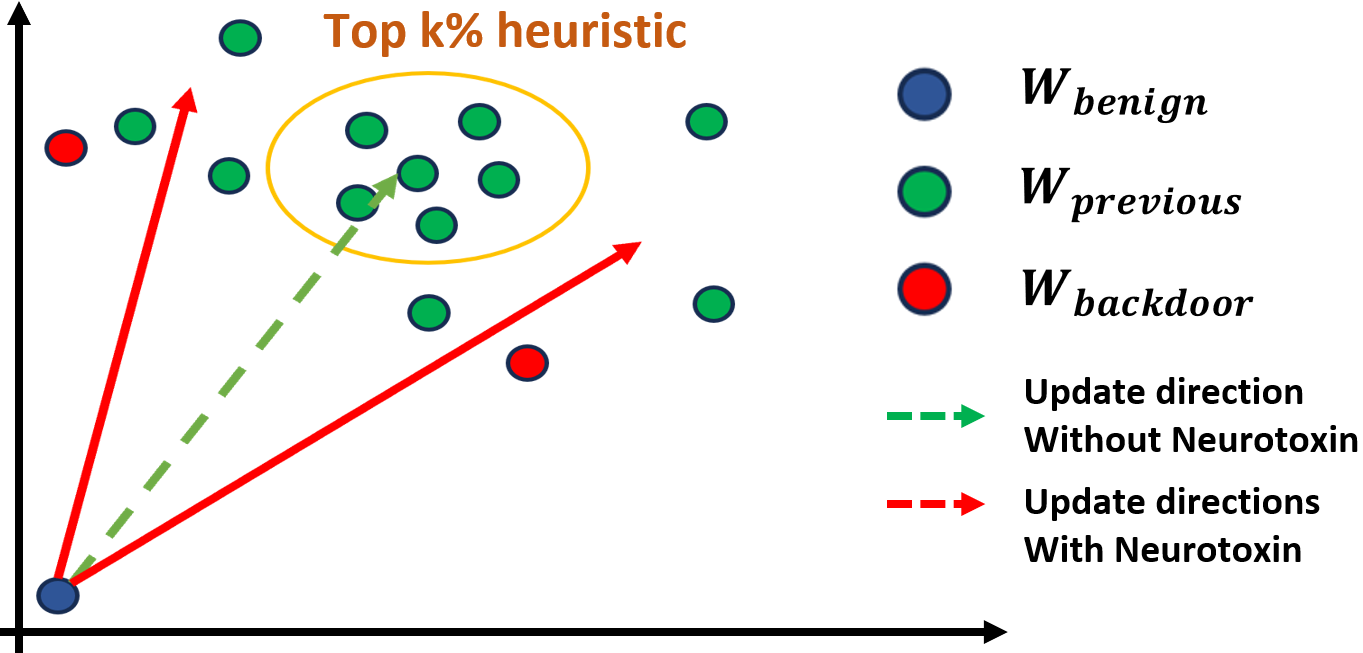}}
\caption{Neurotoxin.}
\label{neurotoxinfig}
\vspace{-20pt}
\end{figure}


    
\noindent\textbf{Defense.}
We further conduct multiple defense strategies to demonstrate the superiority of \name. Specifically, we use differential privacy~\cite{xie2021crfl}, model pruning~\cite{liu2018fine}, FLAME~\cite{nguyen2022flame} and SparseFed~\cite{panda2022sparsefed}  to resist backdoor attacks. Figure~\ref{defense} shows that \name can still maintain high effectiveness under multiple defense strategies. Taking the label flipping on CIFAR-10 as an example, it can be observed that our method persistently maintains a high attack success rate even under defense strategies, while other baselines fail.

\noindent\textbf{Analysis on Neurotoxin.} \label{neuro}
Neurotoxin aims to address the durability issue of backdoor attacks in FL settings, which ensures that the backdoor persists in the global model even after the attacker stops uploading poisoned updates. During FL training, Neurotoxin leverages the concept of the L2 norm, a mathematical function that represents the magnitude of a vector. Neurotoxin observes that the majority of the L2 norm of the aggregated benign gradient is contained in a small number of coordinates, which implies that benign updates tend to cluster in a narrow range and consequently, the aggregated gradient direction is likely to point towards this cluster. Therefore, Neurotoxin identifies and targets the parameters that get minimal changes in magnitude during training, as illustrated in Figure \ref{neurotoxinfig} in the Appendix. These relatively stable parameters are less likely to be significantly affected by benign updates, thereby mitigating the risk of the backdoor being overwritten by future updates.

However, our results do not confirm Neurotoxin's expected effectiveness. Despite using this strategy, backdoor accuracy declines similarly to the baseline attack after the attacker exits at round 50. This trend is also observed in \name experiments, warranting further investigation.




\section{Conclusion}

This paper introduces \name, a novel backdoor attack specifically tailored for non-IID federated learning environments utilizing the Scaffold aggregation algorithm. By employing a GAN-based data augmentation technique and exploiting the Scaffold's control variate, \name achieves superior effectiveness, stealthiness, and durability compared to existing methods. Our experimental results on multiple benchmark datasets demonstrate the attack's effectiveness, with the backdoor persisting significantly longer than existing approaches. In the future, we hope researchers can design more robust defense mechanisms to safeguard federated learning systems against such attacks.

{
    \small
    \bibliographystyle{ieeenat_fullname}
    \bibliography{main}
}

\end{document}